\documentclass[runningheads]{llncs}
\usepackage{graphicx}
\usepackage{enumitem}
\usepackage{graphicx}
\usepackage{subcaption}
\usepackage{amsmath,amssymb}
\usepackage{color}
\usepackage{fancyhdr}
\usepackage{hyperref}

\hypersetup{
    colorlinks=true,
    linkcolor=blue,
    filecolor=magenta,      
    urlcolor=cyan,
}

\begin{document}
\renewcommand{\headrulewidth}{0pt}
\mainmatter

\def\ACCV20SubNumber{542}  

\title{Encode the Unseen: Predictive Video Hashing for Scalable Mid-Stream Retrieval} 
\titlerunning{Predictive Video Hashing for Scalable Mid-Stream Retrieval}
%
\author{Tong Yu\and
Nicolas Padoy}
\authorrunning{Tong Yu \and Nicolas Padoy}
%
\institute{ICube, University of Strasbourg, CNRS, IHU Strasbourg, France \\
\email{tyu@unistra.fr, npadoy@unistra.fr}
}

\maketitle

\fancypagestyle{firststyle}
{
   \fancyhf{}
   \rfoot{\textbf{Accepted at ACCV 2020}}
}

\thispagestyle{firststyle}

\begin{figure}
    \centering
    \includegraphics[width=1.0\linewidth]{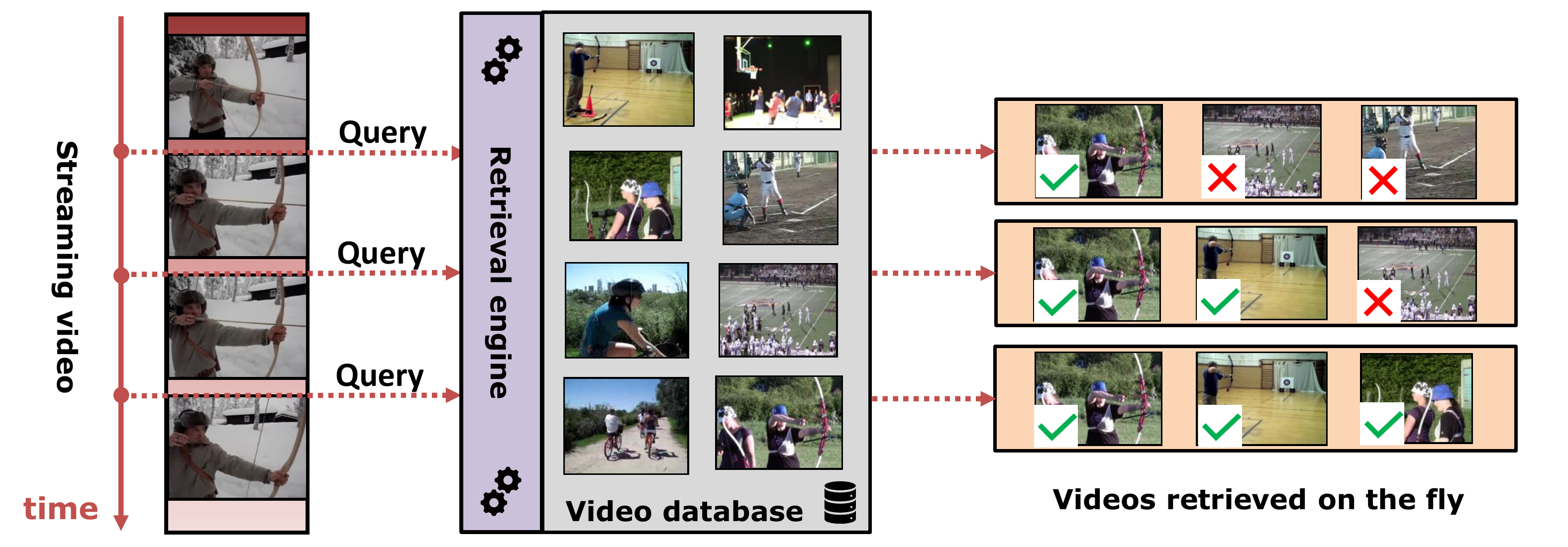}
    \caption{\small{Our hashing technique enables incremental video retrieval from incomplete queries, as needed for real-time retrieval in live video streams. A streamed video of archery is used for querying. On the last query all 3 returned videos are relevant.}}
    \label{fig:generalintro}
\end{figure}%
\begin{abstract}
This paper tackles a new problem in computer vision: mid-stream video-to-video retrieval. This task, which consists in searching a database for content similar to a video right as it is playing, e.g. from a live stream, exhibits challenging characteristics. Only the beginning part of the video is available as query and new frames are constantly added as the video plays out. To perform retrieval in this demanding situation, we propose an approach based on a binary encoder that is both \textbf{predictive} and \textbf{incremental} in order to (1) account for the missing video content at query time and (2) keep up with repeated, continuously evolving queries throughout the streaming. In particular, we present the first hashing framework that infers the unseen future content of a currently playing video. Experiments on FCVID and ActivityNet demonstrate the feasibility of this task. Our approach  also yields a significant mAP@20 performance increase compared to a baseline adapted from the literature for this task, for instance 7.4\% (2.6\%) increase at 20\% (50\%) of elapsed runtime on FCVID using bitcodes of size 192 bits.
\end{abstract}

\section{Introduction}

\textbf{Video-to-video retrieval} is an emerging problem in computer vision due to the growth of large-scale video databases such as Youtube (720K hours uploaded daily) or Instagram. Current search modalities for those databases rely on keywords, which are effective to some degree for an adequately tagged set of videos but inevitably fail to capture the rich visual information contained in video data.

Solutions for visual search are already widely in use for static images, with publicly available services such as TinEye or Google Image Search. These tools take an image supplied by the user and return a list of visually similar pictures, allowing the user to search for content that would otherwise be difficult to find via a text-based search engine. Video, on the other hand, has not received nearly the same amount of attention when it comes to retrieval tasks. Furthermore, the few existing works on video retrieval treat the problem as a pure postprocessing task by only considering prerecorded videos as inputs. There is, however, huge potential for retrieving content directly from non-prerecorded \textbf{live video sources}, in order to provide \textbf{dynamic search results}. Recent advances in text-based search engines, which can now refresh results while the user is typing, prove how useful this innovation would be.
This work therefore introduces a framework for \textbf{unsupervised real-time-stream-to-video retrieval} (Figure \ref{fig:generalintro}) and proposes, to our best knowledge, the first solution to this problem. Popular livestreaming platforms such as Twitch or Youtube Live would highly benefit from such a method, which dynamically searches for content from the live stream. This would enable on-the-fly content suggestion, a powerful feature with a wide variety of potential applications:

\begin{itemize}
    \setlength\itemsep{0.05em}
    \item \textbf{advertising}: commercials, content curation relevant to a live stream
    \item \textbf{sports \& entertainment}: highlights during live events, enhanced by replays of similar past content
    \item \textbf{education}: live tutoring. A person films themself cooking or performing repairs, while an algorithm follows and assists their process in real time with retrieved instructional clips
    \item \textbf{surveillance \& security}: anomaly detection from a live security feed. An algorithm continuously searches a database of security events to provide reference for monitoring the feed in real time
    \item \textbf{healthcare}: relevant surgical actions shown to a surgeon simultaneous to real-time events in the intervention
\end{itemize}

This task presents two main challenges that make it particularly difficult: 

\begin{enumerate}[label=\Alph*.]
    \item The search protocol needs to be \textbf{fast} enough for real-time use. This is a brutal constraint, considering the typical size of a large-scale video database, as well as the size of video data itself compared to static image data.
    \item The real-time streaming setting implies that the video is \textbf{incomplete}; at the beginning of a video, most of the visual information is missing.
\end{enumerate}

An efficient solution for technical challenge A is \textbf{hashing} \cite{ssth,ssvh,udvhlstm,udvhtsn,jtae,nph}. Hashing employs a hashing function $\phi$ that maps high dimensional data entries - in our case videos - to compact binary codes, or \textbf{bitcodes}. A good choice of $\phi$ should respect visual similarity: two videos with similar visual content should yield bitcodes differing only by a small number of bits. Once $\phi$ is determined, the video database to search from is mapped to a set of bitcodes referred to as the \textbf{codebook} (also called \textit{gallery} in certain papers). Search can then be performed in the following manner: the query video is mapped to its bitcode, the \textbf{hamming distance} between the query's bitcode and each bitcode in the codebook is computed, and finally the top K closest entries are returned. This search protocol can be extremely fast: distance computation has linear complexity w.r.t. the number codebook entries, and each distance computation can be performed as a bitwise XOR operation. For mid-stream retrieval however, an additional highly desirable feature is \textbf{incrementality}: new information added by the currently playing video should be incorporated into the bitcode without reprocessing the entire video from its beginning. This keeps time and space requirements to a minimum, which is crucial for scalability when considering extremely large video databases, as well as large numbers of simultaneous queries to process.

Our first contribution is therefore to propose \textbf{the first incremental video hashing method}. It relies on a self-supervised binary video encoder consisting of a 3D CNN combined with a RNN. RNNs are naturally suitable for incremental tasks; their output at each timestep only depends on the corresponding new input and the value of their internal memory, which accounts for all previous timesteps on its own. Existing video hashing approaches, which did not consider the case of partial videos, either use pooling \cite{udvhtsn}, or employ RNNs with sampling schemes that render them non-incremental \cite{ssth,ssvh,udvhlstm,jtae,nph}. The expected use case for those methods is a single retrieval from a full video taken from beginning to end, while our method generates bitcodes throughout the video's duration, making it capable of supporting repeated mid-stream queries.

In order to deal with the incomplete videos, our approach needs to learn to anticipate the future video content. We therefore introduce our second contribution, which is \textbf{the first predictive hashing method}. Instead of generating bitcodes according to what is actually seen, which is the paradigm followed by all image \cite{itq,sphhash,distillhash,knnhash} and video \cite{ssth,ssvh,udvhlstm,jtae,nph} hashing methods up to this point, we generate them according to what is \textit{expected}. To achieve this, we introduce two unsupervised knowledge distillation methods where the binary encoder, by learning to anticipate future content from incomplete videos, generates bitcodes that account for the unseen part of the video in addition to the part it has already seen. This form of distillation relies on \textit{binary} embeddings, and is radically different from other methods considering future frames \cite{teacherstudent,earlysurgery,joint_anticip}, as these have employed \textit{real-valued} embeddings and targeted tasks completely different from retrieval. The enriched bitcodes formed by our method produce better matches with relevant content in the database and significantly improve retrieval results when incomplete queries are used.

\section{Related work}
\subsection{Video activity understanding}
Due to the high dimensionality of the data involved, deep neural networks have become the go-to approach in video activity understanding methods. Videos may be considered as sets of static frames with each frame processed individually -for instance with a 2D CNN. However, a video-specific deep architecture should account for temporal dependencies in video data. A wide range of methods were designed for this purpose; we mention a few below that are relevant to this work.

The LRCN architecture \cite{lrcn} is an early example of temporal and spatial model combination. A CNN maps video frames to low-dimensional visual feature vectors. Those features are then fed to an LSTM \cite{lstm}, responsible for learning the video's temporal dynamics in order to perform tasks such as video captioning.
Similarly, \cite{videoautoencoder} runs an LSTM autoencoder on VGG features. Trained in an unsupervised manner, this model learns compact and discriminative representations of videos that can be repurposed for action recognition tasks.
Two-stream networks \cite{twostream1,twostream2,twostream3} approach the problem from a different angle, by combining two 2D CNNs: one for color data and one for optical flow.
Instead of accounting for the spatial aspect and the temporal aspect of the problem in two separate components, the TSM \cite{tsm} and 3D CNN approaches \cite{c3d,i3d,r3d1,r3d2,slowfast} combine them into a single architecture. The C3D model\cite{c3d}, and later on the I3D model \cite{i3d} employ 3D convolutions (one temporal and two spatial dimensions) in order to directly process video inputs. I3D is the architecture we use for visual feature extraction.

\subsection{Early activity recognition}
In early activity recognition (EAR) tasks, a model attempts to classify an action depicted in a video by only observing its beginning. The potentially large amount of missing visual information introduces a level of difficulty that is not found in general activity recognition tasks, but which we also encounter in our problem.

Methods such as \cite{earlyloss} and \cite{msrnn} rely on elapsed time or progress. \cite{earlyloss} trains an LSTM model with a temporally weighted loss enhancing the importance of predictions made early on in the video. \cite{msrnn} proposes a method for early recognition based on soft labels at different levels of video progress. A different type of approach for early recognition tasks is to synthesize the content in the video's future: \cite{deepscn} attempts to guess future frame embeddings in videos with a multi-layer fully connected network. \cite{joint_anticip} employs video representations fed to generator models trained to synthesize embeddings future frame embeddings. Finally, teacher-student distillation methods train a student model that has not seen future frames to replicate representations learnt by a teacher model that has. \cite{earlysurgery} employs a distillation method that trains an LSTM to predict the state of another LSTM at a further point in the input video. A different distillation method, as introduced in \cite{teacherstudent}, trains a forward-directional LSTM to match at each timestep the hidden states of a trained bidirectional LSTM.

Although EAR and mid-stream video retrieval operate on similar inputs, which are the early parts of videos, EAR is entirely different from our problem in the very same sense that recognition is different from retrieval. Retrieval involves comparisons to all items in a database to retrieve those that are similar. Similarity in this context can consist of much more than simply having identical class labels, although the evaluation protocol relies on class labels to provide quantitative results. EAR, on the other hand, is entirely label-dependent. The output of EAR is a class, while the output of our approach consists of a large selection of items from an entire video database.

\subsection{Video hashing}
\label{section:relwork:videohashing}
For static image data, a large number of hashing methods exist in the literature \cite{itq,sphhash,distillhash,knnhash}.
Video data is substantially more challenging. First, video data requires capturing \textbf{rich temporal dependencies}. Second, \textbf{the sequence length is variable}: reducing a video to a fixed-length representation is not straightforward. Third, \textbf{data dimensionality is considerably higher}: a minute of video at 30 fps contains 1800 times the amount of data contained in any of its frames. Capturing enough visual information from it in a single vector of a few hundreds, if not tens of bits is a tough challenge. Finally, large-scale video databases require considerably larger amounts of time than images to annotate; \textbf{unsupervised learning} is therefore heavily favored.

Six unsupervised video hashing methods have addressed the first challenge in a significant manner by incorporating some degree of temporal modeling:  SSTH \cite{ssth} uses an encoder-decoder pair of RNNs with a binary state for the encoder in order to learn rich, discriminative bitcodes. SSVH \cite{ssvh} improves over this architecture by aggregating information at different time scales in the video with stacked RNNs. JTAE \cite{jtae} also follows the encoder-decoder type of approach with separate encoders for appearance and temporal dynamics. NPH \cite{nph} uses an encoder-decoder similarly to the previous approaches, but trained with losses that attempt to preserve neighborhood structures from the embeddings it receives. UDVH-LSTM \cite{udvhlstm} applies geometric transformations to embeddings obtained from temporally pooled features from a CNN-LSTM, before binarizing those embeddings. UDVH-TSN \cite{udvhtsn} employs a similar approach, replacing the CNN-LSTM with a modified Temporal Segment Network \cite{tsn}.

All those methods deal with the issue of variable sequence length by resorting to fixed-rate sampling schemes, rendering them non-incremental. Those schemes select a set number of evenly-spaced frames in all input videos and discard the rest. This results in a fixed-length but variable-rate and sparse representation that would be unsuitable in a live setting, as explained in Section \ref{sec:sampling-extraction}. All methods other than SSTH also exhibit additional characteristics that are unsuitable for incremental use in real-time: feature pooling from the beginning to the end of a video \cite{jtae,udvhlstm,udvhtsn}, hierarchical aggregation of timesteps \cite{ssvh}, video anchor fetching at query time \cite{nph}. For this reason, we build our baseline approach on SSTH.
\textbf{None of the aforementioned approaches have examined the mid-stream, incomplete video case.}

\section{Methods}
\subsection{Overview}
The problem can be formally stated as follows: given a \textbf{level of observation} $\alpha \in  (0, 1]$, a video $\mathcal{Q}$ with duration $T(\mathcal{Q})$, a query $\mathcal{Q}_{\alpha}$ consisting of $\mathcal{Q}$ truncated mid-stream at time $\alpha T(\mathcal{Q})$, and a database of videos $\mathcal{B}=\{V_{0} ... V_{N}\}$, use only $\mathcal{Q}_{\alpha}$ to find the $K$ videos from $\mathcal{B}$ most similar to $\mathcal{Q}$.

Since hashing is the method employed here, $\mathcal{B}$ and $\mathcal{Q}$ are represented by binary codes during search operations. The overall protocol is therefore composed of the following steps: video-level feature extraction, self-supervised hash function training, codebook mapping and finally query operations.

\subsection{Sampling \& feature extraction}
\label{sec:sampling-extraction}

As mentioned in the related work section, previous approaches have chosen sparse, fixed-length video-level representations which can result in the loss of large amounts of visual information in long videos. More importantly, such sampling and feature extraction schemes are inadequate for real-time usage: the sampling rate in that case is dependent on the length of the full video, which would be unknown mid-stream. An early retrieval algorithm should be capable of delivering search results at a fast and steady pace.

To solve this issue, we have chosen a dense, fixed-rate sampling and feature extraction scheme that leverages a 3D CNN to aggregate visual information over short time windows. The chosen architecture is I3D \cite{i3d}, pretrained on the Kinetics \cite{kinetics} dataset for action recognition. Input frames from the videos are center-cropped to $224 \times 224$. Clip input length is 64 frames (roughly 2 seconds at 30 fps). This yields a feature tensor with dimensions $8 \times 7 \times 7 \times 1024$, which we average pool into a feature vector of size $N_{f} =  1 \times 2 \times 2 \times 1024 = 4096$. The entire video is processed in consecutive clips, resulting in a 30 $/$ 64 $=$ 0.47 Hz constant feature output rate.

\subsection{Binary RNN encoder}

\begin{figure}[t]
\begin{center}
   \includegraphics[width=0.8\linewidth]{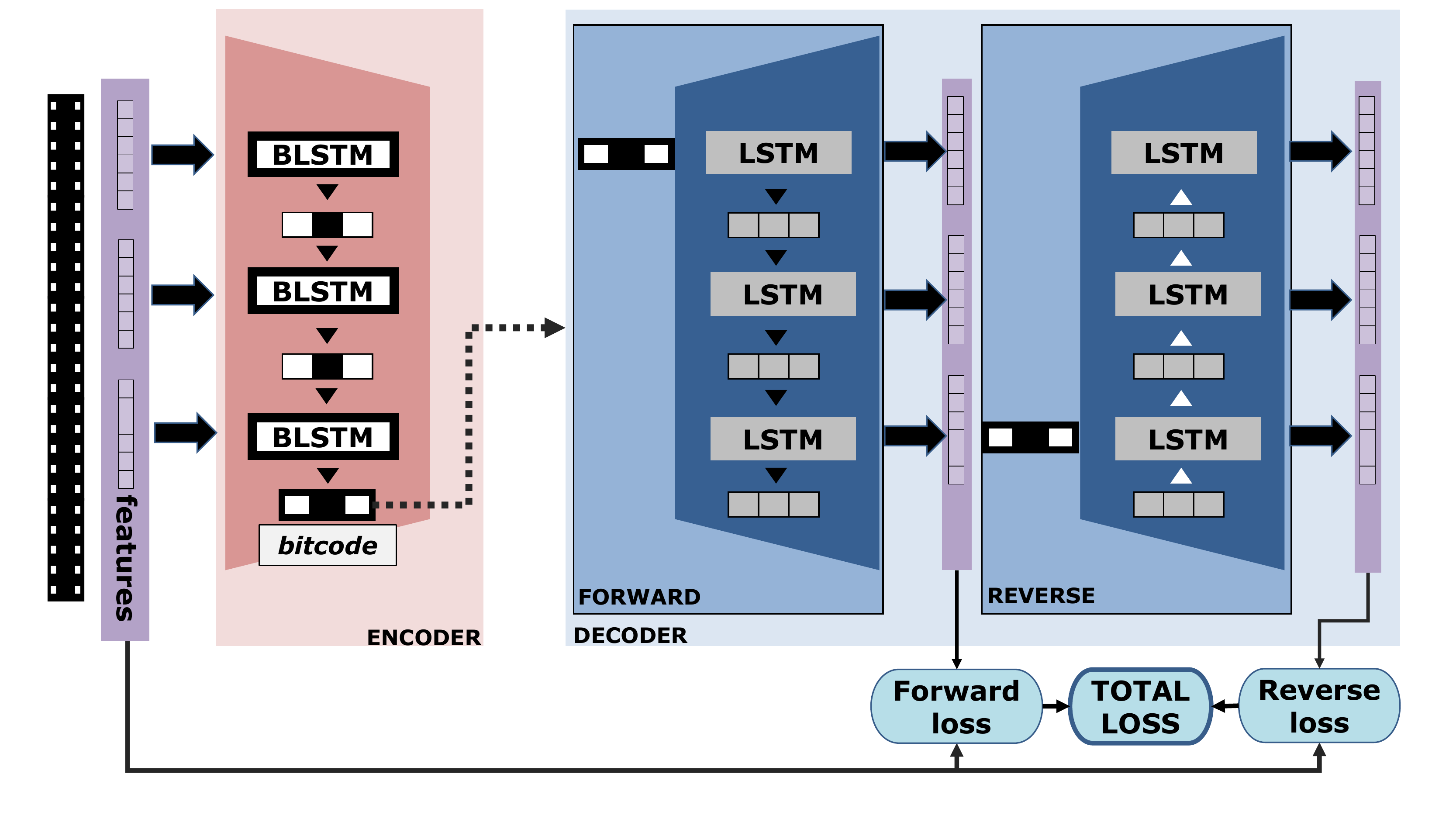}
\end{center}
   \caption{\small{Illustration of the self-supervised binary video autoencoder \cite{ssth}.}}
\label{fig:binary_autoencoder}
\label{fig:onecol}
\end{figure}

An RNN (Figure \ref{fig:binary_autoencoder}) takes the feature vectors $f_{0},f_{1},...,f_{t},...,f_{T(V)}$ from I3D in chronological order and maps each one to a bitcode of size $N_{bits}$. This RNN is built on two stacked LSTMs, with a binary LSTM as introduced in \cite{ssth}. The hidden state circulating from one RNN timestep $t$ to the next is the above mentioned bitcode noted $b(V, t)$ for a video $V$. Batch normalization is applied to the cell state as done in \cite{ssth}. This RNN is trained as the encoder part of an autoencoder model. The final hidden state of the decoder - the video's bitcode - is passed to two decoder RNNs tasked with reconstructing the sequence of input feature vectors in opposite orders. This training process encourages the encoder to incorporate rich, discriminative information into the bitcodes.

Formally, the autoencoder is trained to minimize the following \textbf{unsupervised} loss function: $\mathcal{L}_{decoder}(V) = \sum_{j=0}^{T(V)}  \|f_{j} - \overleftarrow{f}_{j}\|_{2} + \|f_{j} - \overrightarrow{f}_{j}\|_{2}$

with $\overleftarrow{f}, \overrightarrow{f}$ being the reverse and forward reconstructed sequences, respectively.

\begin{figure*}[t]
\begin{center}
   \includegraphics[width=1.0\linewidth]{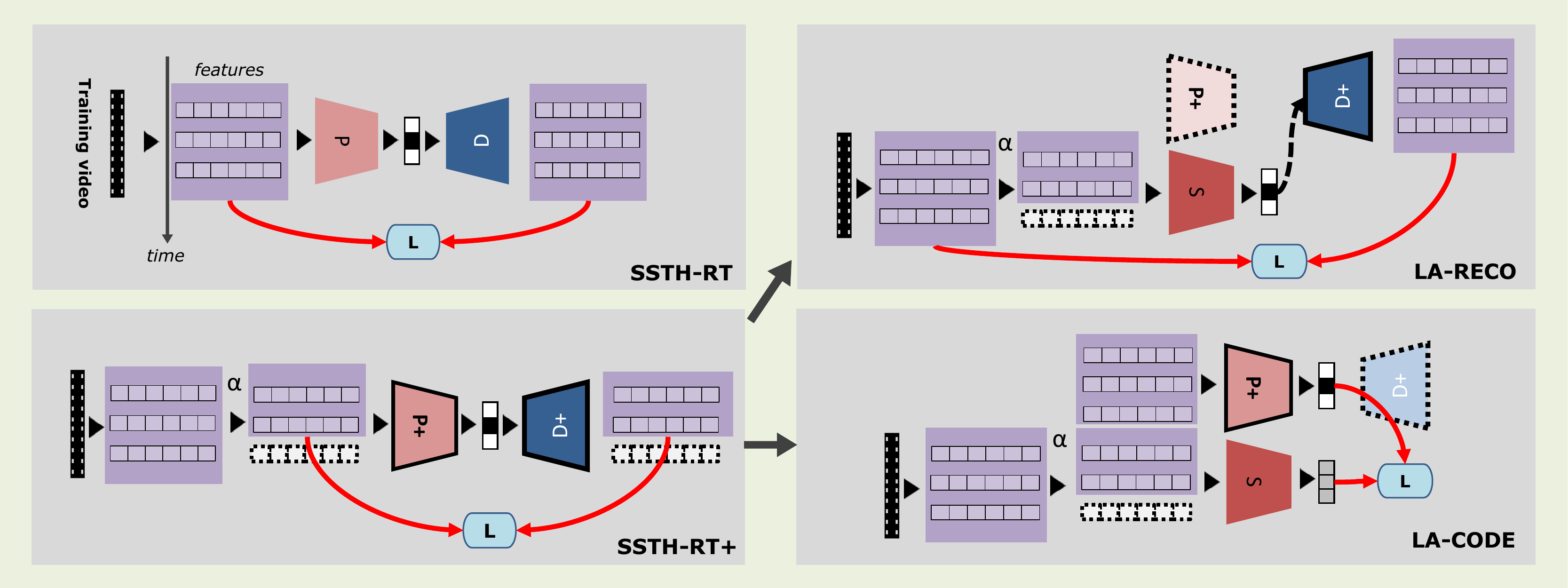}
\end{center}
   \caption{\small{Overview of training setups for all models. P stands for primary encoder, S for secondary encoder, D for decoder and L for loss. The (+) indicates the use of truncated videos during training. Elements with dashed contours are unused.}}
\label{fig:training_setup}
\end{figure*}

Backpropagation relies on the gradient substitution method shown in \cite{ssth}. Trained as such with full videos, we call this configuration \textbf{SSTH-RT} (Figure \ref{fig:training_setup}, top left) for real-time, as each new increment of the video playing in real-time is incorporated into the bitcode. This is, in the form presented here, our first baseline approach.

\subsection{Data-augmentated encoder via truncated training duplicates}
\begin{figure*}[t]
\begin{center}
   \includegraphics[width=1.0\linewidth]{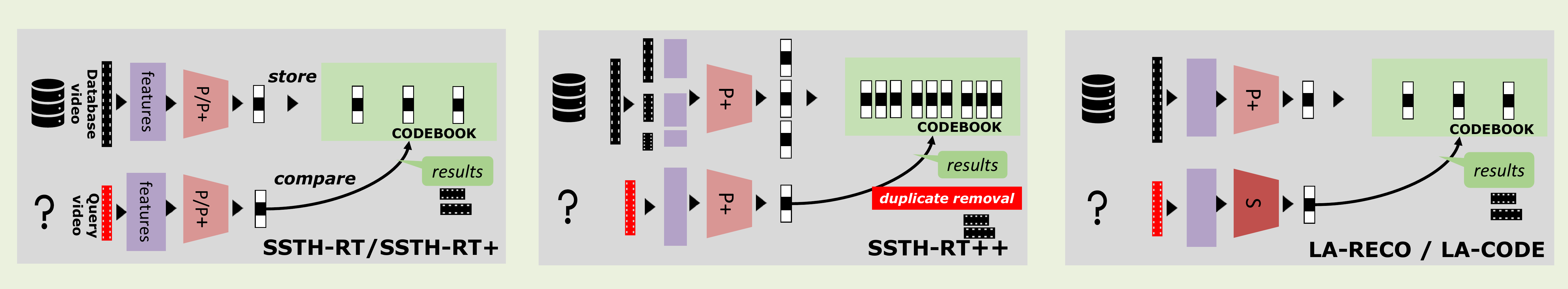}
\end{center}
   \caption{\small{Overview of retrieval setups for all models, depicting the generation of the codebooks, their size, as well as the trained encoders used.}}
\label{fig:eval_setup}
\end{figure*}

Since the training process for the baseline setting always involves complete videos, a straightforward data augmentation method for training the autoencoder is to simply give the autoencoder videos truncated at various levels of observation to reconstruct (Figure \ref{fig:eval_setup}, left). This approach provides some training exposure to videos that are incomplete. This is referred to as \textbf{SSTH-RT$^{+}$} in the later sections.

\subsection{Augmented codebook with truncated database duplicates}
Another idea to improve early retrieval results is to incorporate trimmed videos, both during training - as in SSTH-RT$^{+}$ - and codebook construction (Figure \ref{fig:eval_setup}, center). We refer to this later on as \textbf{SSTH-RT$^{++}$}. While this idea might seem appealing at a first glance, it causes the search protocol to slow down due to the insertion of bitcodes from trimmed duplicates. This requires inserting as many duplicates per video as levels of observation employed - in our case: 10 from 0.1 to 1.0 - which is also a source of scalability issues. Assuming $N_{\alpha}$ levels of observation are employed, search time and storage space requirements both get multiplied by $N_{\alpha}$. Duplicates in the search results, namely, videos truncated at different levels of observations from the same original video, would also need to be purged, the cost of which has $N_{cb} \times N_{\alpha}$ complexity, $N_{cb}$ being the size of the codebook without duplicates. Correctly retrieved results that are duplicates may improve retrieval metrics, but do not suit general application cases: a search operation returning truncated versions of the same video would be unhelpful to the user. The approaches proposed in the next section are faster as they employ codebooks with the exact same population size as the original video database.

\subsection{Look-ahead distillation for encoders}
We propose two forms of distillation in order to train \textbf{predictive encoders} that can anticipate the future content when presented with incomplete videos, and therefore generate richer, more discriminative bitcodes that are more likely to yield better search results.

\subsubsection{Indirect distillation: look-ahead reconstruction (LA-RECO) loss }
An encoder, which will from now on be referred to as the primary encoder (P in Figure \ref{fig:training_setup}), is trained following the same process as SSTH-RT$^{+}$ jointly with a decoder. Once this training is done, the main encoder is set aside and a secondary encoder (S in Figure \ref{fig:training_setup}, top right) is introduced. Only incomplete videos are fed into it - i.e. videos truncated to $\alpha T(V)$ with randomized $alpha$ for each training step. The resulting bitcode is passed to the trained decoder with frozen parameters. The output of the decoder is compared to the full sequence in the loss, not just the truncated part fed to the secondary. This forces the secondary encoder to guess a representation accounting for future video frames.

\subsubsection{Direct distillation: look-ahead bitcode (LA-CODE) loss}
A secondary encoder is introduced as in the previous section, however this time the primary is put to use and the decoder is set aside (Figure \ref{fig:training_setup}, bottom right). During training, the primary encoder, which has its parameters frozen, receives the entire video while the secondary encoder is only fed the $\alpha T$ first frames, again with randomized $alpha$. From the secondary, we extract the real-valued input $\beta$ to the \textit{sgn} function that leads to the last bitcode - we refer to this as the \textit{prebitcode} - and compare that to the bitcode $b$ given by the primary using an L2 loss function:

\begin{equation}
    \mathcal{L}_{LA-CODE} = \| \beta(V, \alpha T) - b(V) \|_{2}.
\end{equation}

This conditions the encoder, despite being only fed part of a video, to mimic the full video's bitcode. 

\subsection{Experimental setup}

\subsubsection{Datasets}
For our experiments we employed two datasets. The first is FCVID \cite{fcvid}, a public video dataset used for research on video retrieval and activity understanding depicting various activities, objects, sports and events. 234 video classes are present. The average video duration is $134 \pm 92$ seconds. Data was split as follows: 45K for training, 45K for testing. Within the test videos, 42.5K constitute the database while the remaining 2.5k (about 10 videos per class on average) are used for querying.
The second dataset we employed is ActivityNet \cite{activitynet}. We used the 18K videos available for download. 200 activity classes are featured. A number of videos had their labels concealed by the organizers of the ActivityNet challenge; those videos were put in the training set. The average duration is $117 \pm 67$ seconds. Data was split as follows: 9K for training, 9K for testing, and within the testing videos 8K for the codebook, 1K (about 5 per class on average) for the query.
In both datasets, the videos are untrimmed, which implies that the content reflecting a given video's class may not be shown for its entire duration. All videos have been resampled at 30 fps and a few excessively long videos have been cropped to 4 min 30s.

\subsubsection{Training}
Video-level features are extracted separately by I3D and reloaded from binary files. Training is done in batches; to achieve batching with variable sequence lengths we grouped videos of similar length into buckets and randomly pick batches from a single bucket at a time during training. Batches are then trimmed to the shortest duration present in each of them.
Encoder/decoder pairs are trained for 60 epochs with a learning rate of $5.10^{-3}$ and a batch size of 40. With $N_{bits}$ being the size of the bitcode, the encoder is a 2-layer RNN with $2 \cdot N_{bits}$ in the first layer and $N_{bits}$ units in the second. The decoder also has two layers, with $N_{bits}$ in the first layer and $2 \cdot N_{bits}$ units in the second. Secondary encoders in LA-RECO or LA-CODE are trained for 15 epochs with a learning rate of $5.10^{-4}$. Their weights are initialized from the values of the trained primary encoder's weights.

\begin{figure}[t]
\begin{center}
  \includegraphics[width=1.0\linewidth]{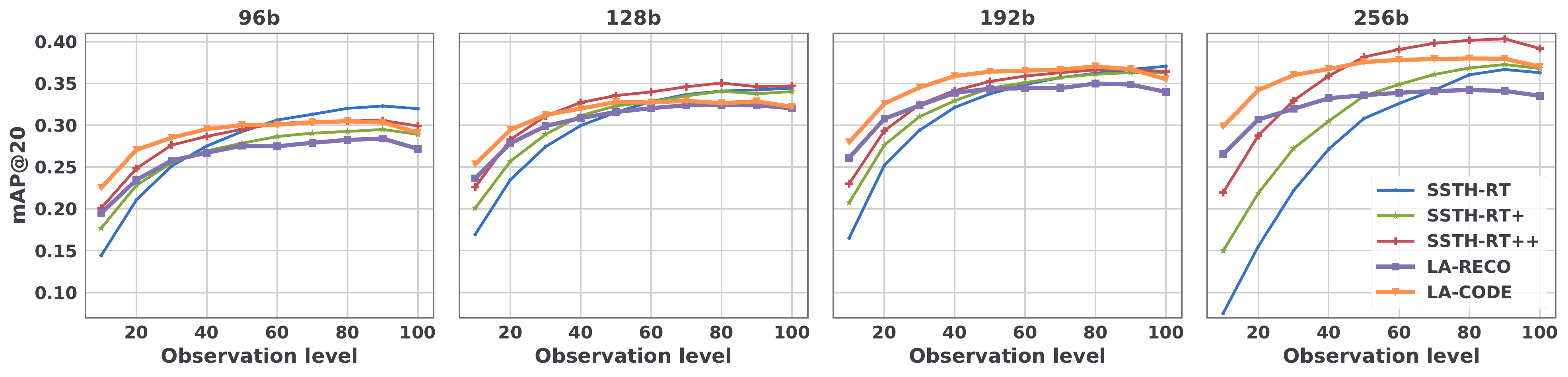}
\end{center}
  \caption{\small{Overall results, \textbf{FCVID}: mAP@20 for all models and 4 different bitcode sizes at 10 different observation levels (\%).}}
\label{fig:main_results}
\end{figure}

\begin{figure}[t]
\begin{center}
  \includegraphics[width=1.0\linewidth]{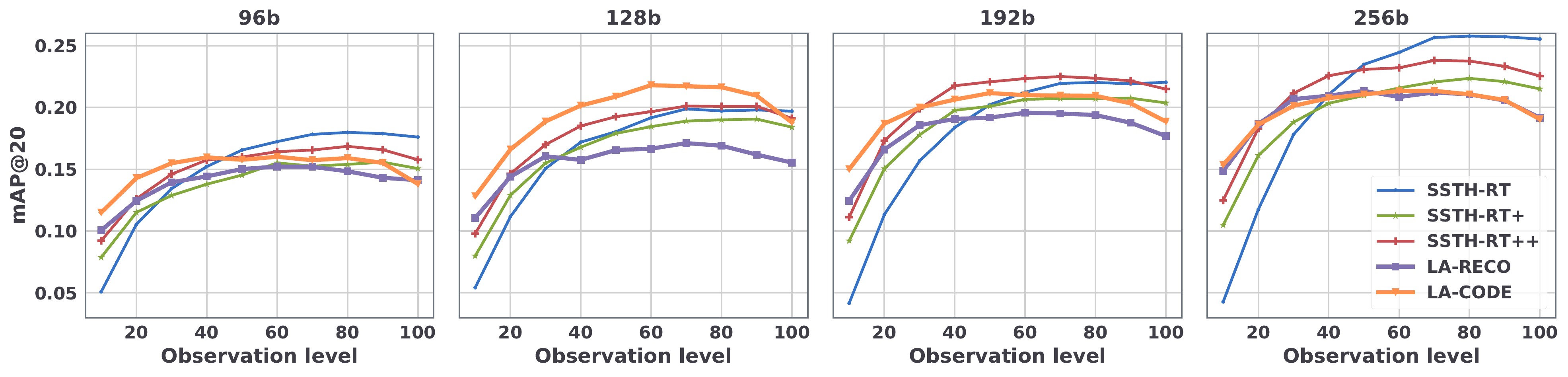}
\end{center}
  \caption{\small{Overall results, \textbf{ActivityNet}: mAP@20 for all models and 4 different bitcode sizes at 10 different observation levels (\%).}}
\label{fig:main_results_an}
\end{figure}

\begin{figure}[t]
\begin{center}
  \includegraphics[width=1.0\linewidth]{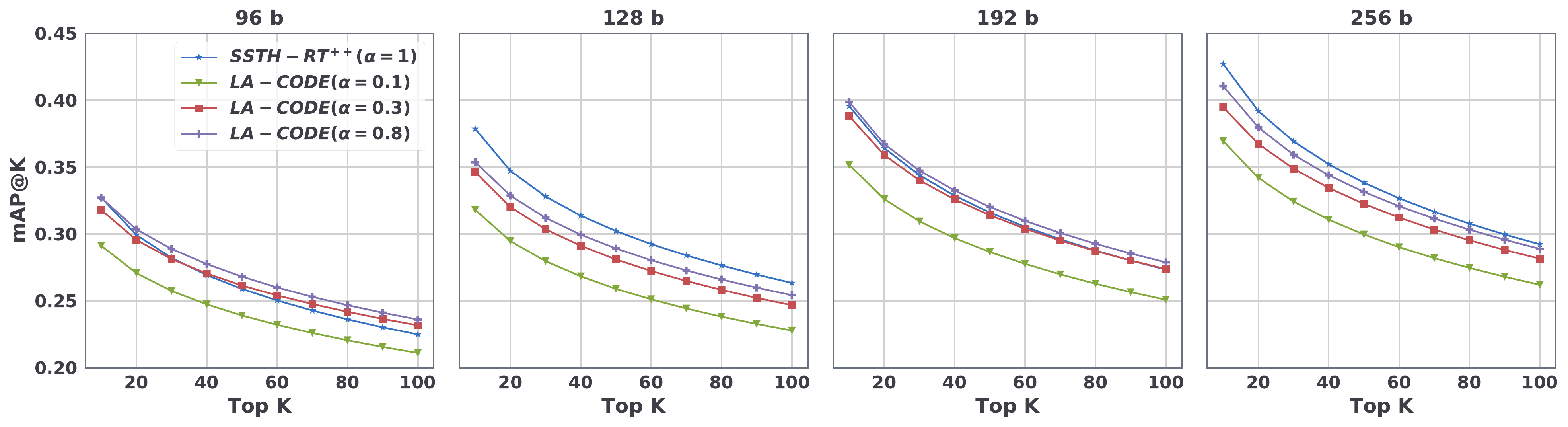}
\end{center}
  \caption{\small{Detailed results, \textbf{FCVID}: mAP@K for LA-CODE and 4 different bitcode sizes at 3 different observation levels (\%).}}
\label{fig:mapk}
\end{figure}

\begin{figure}[t]
\begin{center}
  \includegraphics[width=1.0\linewidth]{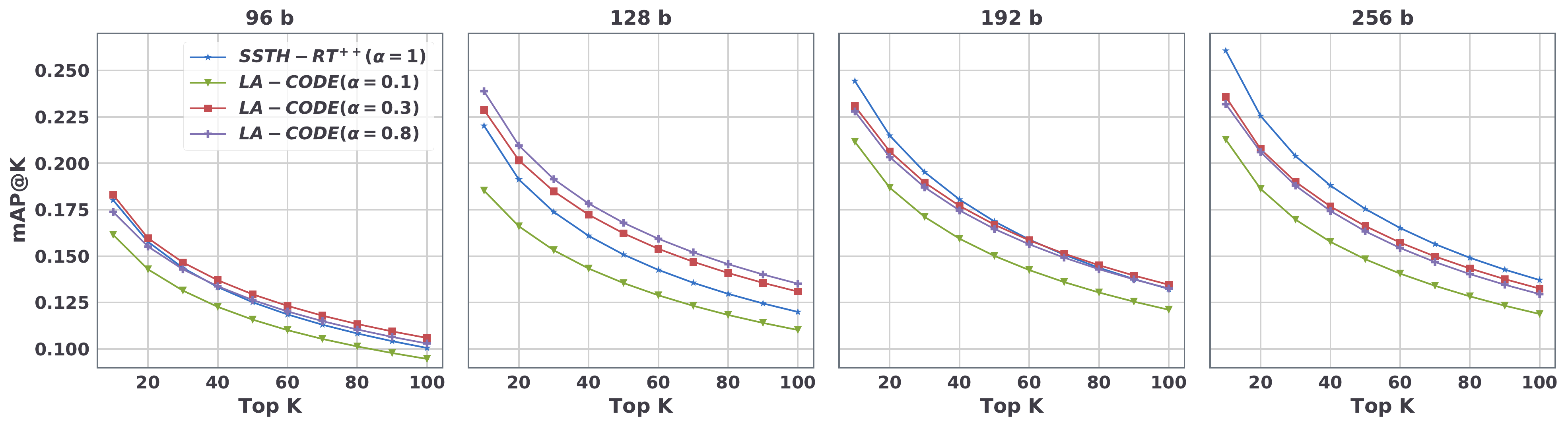}
\end{center}
  \caption{\small{Detailed results, \textbf{ActivityNet}: mAP@K for LA-CODE and 4 different bitcode sizes at 3 different observation levels (\%).}}
\label{fig:mapk_an}
\end{figure}

\subsubsection{Evaluation}
Evaluation is performed as in classical video retrieval by performing queries and comparing them to the search results based on their class labels. This way of assessing visual similarity is limited but also the only available option for quantitative evaluation; qualitative results (Section \ref{qual}) show that information much richer than the classes is captured during retrieval. Once the primary and secondary encoders are trained, we use the primary to generate the codebook from the corresponding videos picked from the test set, as shown in Figure \ref{fig:eval_setup}. The remainder of the test set forms the query set.
The query encoder generates a bitcode for each query video, truncated at $\alpha T$ frames. The code is compared to codebook entries, which are then ranked by Hamming distance to the query bitcode starting from the nearest. Query videos are truncated using 10 different values for $\alpha$ ranging from 0.1 to 1.0, representing different stages of elapsed video.

Average precision at K for a single video, as employed in \cite{ssth}, is given by $AP@K = \frac{1}{K}\sum_{j=0}^{K}\frac{N_{correct}(j)}{j}$, where $N_{correct}(j)$ is the number of matching results among the top $j$.
The mean over all query videos is the Mean Average Precision at K - or mAP@K. We report this metric by level of observation for all methods and all bitcode sizes for K = 20. Lower values of K are more relevant in a real-time use case, where one can only handle a small number of search results; nonetheless we provide mAP@K curves for K up to 100 for our best model. We also provide qualitative retrieval results for two cases. All experiments are repeated for four different bitcode sizes: 96, 128, 192, 256 bits. The compared approaches are:
\begin{itemize}
    \item \textbf{SSTH-RT}: binary autoencoder trained to reconstruct full sequences from full sequences. \textbf{This provides a point of comparison against the literature, since it is SSTH \cite{ssth} implemented with a different feature extraction scheme to cope with fixed-rate sampling}(see Section \ref{sec:sampling-extraction})
    \item \textbf{SSTH-RT$^{+}$}: binary autoencoder trained to reconstruct full sequences from full sequences, as well as partial sequences from partial sequences
    \item \textbf{SSTH-RT$^{++}$}: same encoder as SSTH-RT$^{+}$, with truncated duplicates hashed into the codebook
    \item \textbf{LA-RECO}: SSTH-RT$^{+}$ as primary encoder, with secondary trained to reconstruct full sequences from partial sequences
    \item \textbf{LA-CODE}: SSTH-RT$^{+}$ as primary encoder, with secondary trained to predict the primary's full sequence bitcodes from partial sequences
\end{itemize}

\subsubsection{Implementation details \& performance considerations}
We timed the search protocol in a separate experiment combining all its steps from beginning to end. This includes the I3D forward pass, the encoding, the xor computation with the codebook and the ranking. Results on FCVID show that one search operation using 256b bitcodes and a codebook of 42.5k videos takes 2.7s with a 1080Ti GPU. This enables a query rate of 22 queries per minute, which is sufficient for an end user.

\section{Results}
\begin{table}
\small
\begin{center}
\begin{tabular}{|c||c|c|c||c|c|c||c|c|c||c|c|c|}
\hline
&
\multicolumn{3}{|c||}{96 bits}
&
\multicolumn{3}{|c||}{128 bits}
&
\multicolumn{3}{|c||}{192 bits}
&
\multicolumn{3}{|c|}{256 bits} \\
\hline
FCVID & VE & E & O & VE & E & O & VE & E & O & VE & E & O \\
\hline
SSTH-RT  & 17.8 & 23.5 & 27.6 & 20.2 & 25.9 & 29.9 & 20.9 & 27.4 & 31.8 & 11.5 & 20.6 & 27.9 \\
\hline
SSTH-RT$^{+}$ & 20.3 & 24.2 & 26.6 & 22.9 & 27.6 & 30.6 & 24.2 & 29.4 & 32.6 & 18.5 & 25.6 & 31.0 \\
SSTH-RT$^{++}$ & 22.5 & 26.2 & 28.2 & 25.5 & 29.7 & 32.1 & 26.2 & 30.9 & 33.6 & 25.4 & 31.6 & 35.6 \\
\hline
LA-RECO   & 21.5 & 24.6 & 26.2 & 25.7 & 28.8 & 30.5 & 28.4 & 31.5 & 33.0 & 28.6 & 31.2 & 32.6 \\
LA-CODE   & \textbf{24.8} & \textbf{27.5} & \textbf{28.8} & \textbf{27.4} & \textbf{30.2} & \textbf{31.4} & \textbf{30.3} & \textbf{33.5} & \textbf{35.0} & \textbf{32.0} & \textbf{34.9} & \textbf{36.3} \\
\hline
\hline
ActivityNet & VE & E & O & VE & E & O & VE & E & O & VE & E & O \\
\hline
SSTH-RT  & 7.8 & 12.2 & 14.9 & 8.3 & 13.4 & 16.5 & 7.8 & 14.0 & 17.9 & 8.0 & 15.7 & 20.6 \\
\hline
SSTH-RT$^{+}$ & 9.7 & 12.1 & 13.7 & 10.4 & 14.2 & 16.5 & 12.1 & 16.4 & 18.5 & 13.3 & 17.3 & 19.6 \\
SSTH-RT$^{++}$ & 10.9 & 13.6 & \textbf{15.0} & 12.2 & 15.8 & 17.8 & 14.2 & 18.4 & \textbf{20.3} & 15.4 & \textbf{19.5} & \textbf{21.4} \\
\hline
LA-RECO   & 11.3 & 13.2 & 14.0 & 12.7 & 14.8 & 15.6 & 14.5 & 17.2 & 18.1 & 16.8 & 19.3 & 19.9 \\
LA-CODE   & \textbf{12.9} & \textbf{14.6} & \textbf{15.0} & \textbf{14.7} & \textbf{17.9} & \textbf{19.4} & \textbf{16.8} & \textbf{19.1} & 19.8 & \textbf{17.0} & 19.2 & 19.9 \\
\hline
\end{tabular}
\end{center}
\caption{\small{Results breakdown by $\alpha$ range. For each bitcode size: left column shows very early results (VE, average mAP@20 for $\alpha$ from 0.1 to 0.2); Middle column shows early results (E, average for $\alpha$ from 0.1 to 0.5); Right column shows overall results (O, average over all $\alpha$).}}
\label{table:3ranges}
\end{table}

The mAP@20 results are compared by level of observation for each method in Figure \ref{fig:main_results} with separate graphs for each size of bitcode. SSTH-RT serves as the reference baseline and point of comparison against the literature. SSTH-RT$^{+}$ is a much more competitive model to compare to, SSTH-RT$^{++}$ even more so, but at the cost of 10 times the space requirements and search time.

\textbf{We start with a general overview of the results.}
The general tendency observed is that \textbf{retrieval degrades as the observed portion of video decreases}. With 128 bits and complete videos ($\alpha$ = 1), mAP@20 for SSTH-RT on FCVID reaches 34\% but drops to 30 \% when only the first 40\% of the video are available, down to 17\% with 10\% observation. The trend on ActivityNet is similar with 20\%, 17\% and 5\% mAP@20 respectively.

When considering different bitcode sizes, the tendency observed in Figure \ref{fig:main_results} is that retrieval performance generally increases with size, as more bits enable richer encodings. Looking at one method, bottom mAP@20 for LA-CODE on FCVID starts at 22.5\% for 96 bits, then moves up to 25.3\% for 128 bits, 28.0\% for 192 bits and 29.8\% for 256 bits. In similar fashion, bottom mAP@20 for LA-CODE on ActivityNet respectively reaches 11.4\%, 12.8\%, 15.0\% and  15.4\%.

When considering the different methods, results for non-predictive approaches (SSTH-RT, SSTH-RT$^{+}$ and SSTH-RT$^{++}$) tend to be heavily imbalanced with much higher mAP@20 at high $\alpha$ values. In comparison, results for predictive methods are more even across the $\alpha$ range. An example of this can be seen for 256 bits on ActivityNet, where LA-CODE starts at 30\% and ends at 36\% while SSTH-RT starts at 7\% and ends at 36\%.

\begin{figure*}[t]
\centering
\begin{subfigure}{.48\textwidth}
  \centering
  \includegraphics[width=1\linewidth]{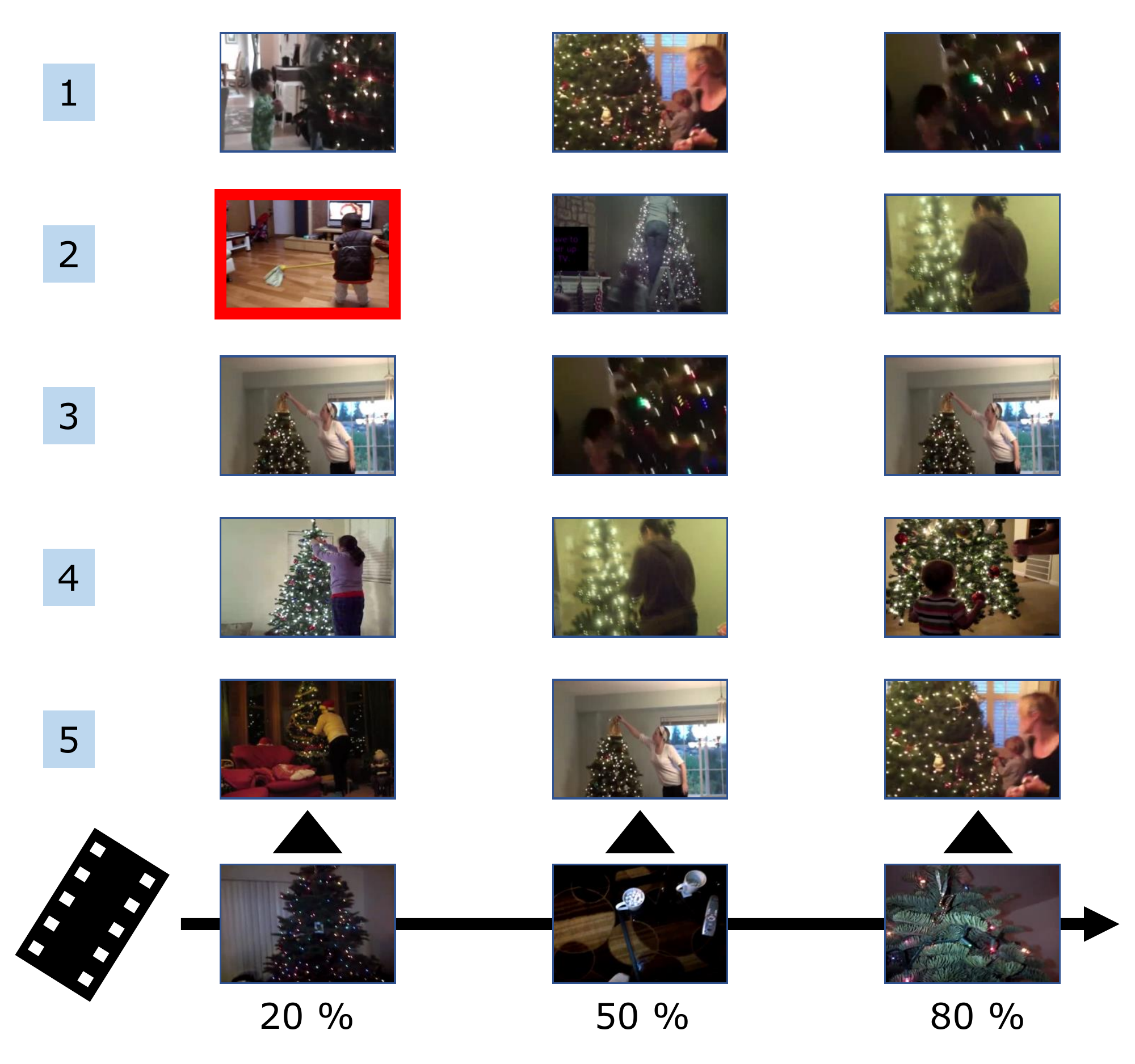}  
  \caption{}
  \label{fig:qual_1}
\end{subfigure}
\begin{subfigure}{.48\textwidth}
  \centering
  \includegraphics[width=1\linewidth]{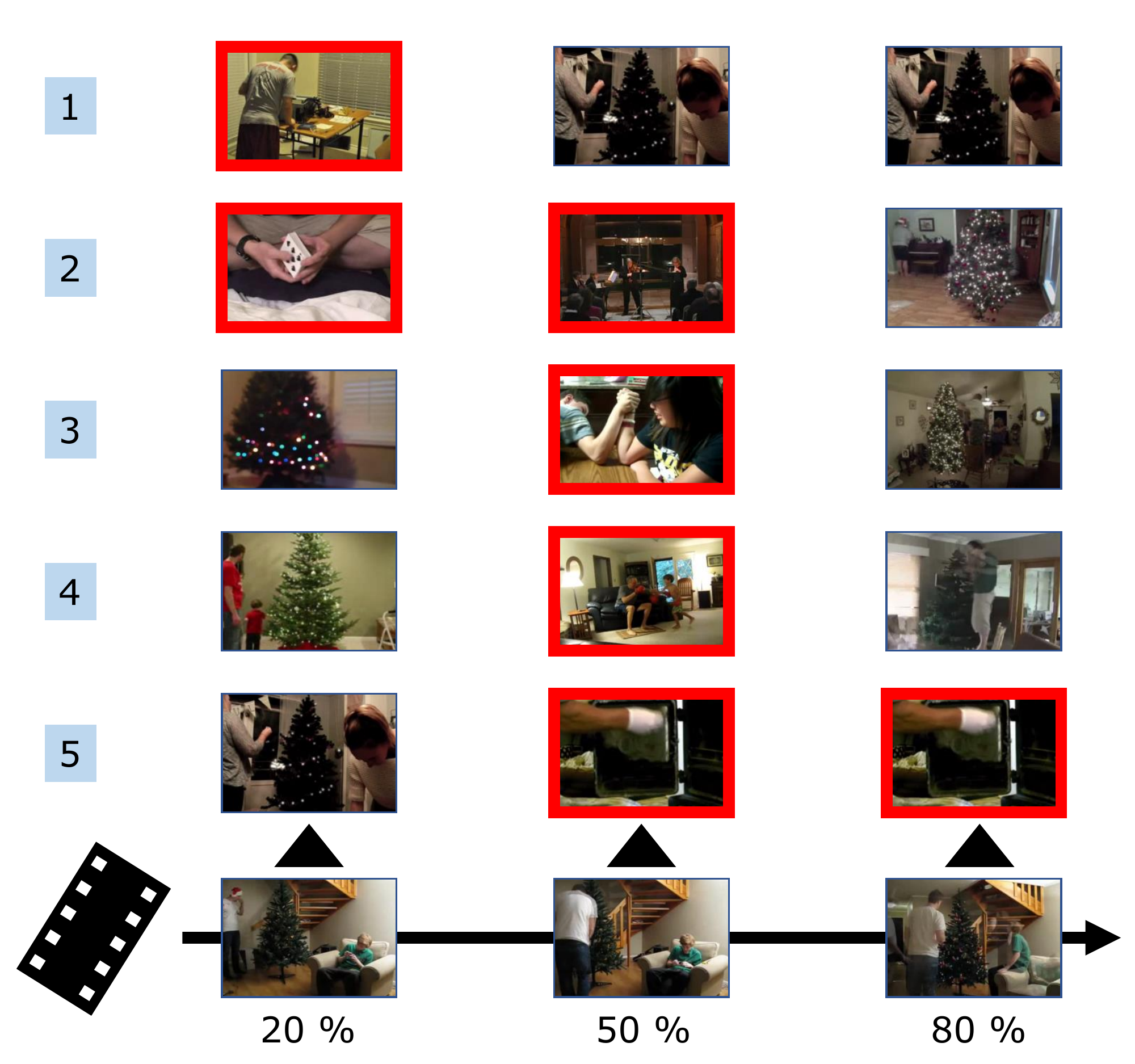}  
  \caption{}
  \label{fig:qual_2}
\end{subfigure}
\caption{\small{Retrieval over time with LA-CODE, illustrating a successful live retrieval (a) and a less successful one (b) on two videos depicting the action {\it decorating a christmas tree}. Red borders indicate a mismatch.}}
\label{fig:qual}
\end{figure*}

\textbf{To allow for clearer comparison, we break down our results into ranges of observation levels.} We consider the average of mAP@20 over very early observations (VE, $\alpha$ from 0.1 to 0.2), early (E, $\alpha$ from 0.1 to 0.5) and overall (O, $\alpha$ from 0.1 to 1.0) in Table \ref{table:3ranges}.

We first consider all levels of observation together, (O) in Table \ref{table:3ranges}. On ActivityNet for 96, 192 and 256 bits SSTH-RT$^{++}$ ranks best overall. This comes, however, at the cost of a codebook 10 times larger and 10 times longer search time, which puts it at a severe disadvantage compared to LA-CODE. For all bitcode sizes on FCVID, for 96 and 128 bits on ActivityNet, LA-CODE is however the top ranking approach.

Although sustaining good performance at higher levels of observation is valuable due to the current level of observation being potentially unknown during streaming in a real use case, \textbf{the focus of this work is on lower levels of observation} (under 50\%). In this range, the main novel challenge specific to our problem, namely missing visual information, is the most manifest.

The performance increase in the very low $\alpha$ range is generally accentuated compared to the whole range of $\alpha$. This makes sense since the losses employed emphasize anticipation. For 192 bits on FCVID, LA-RECO surpasses SSTH-RT$^{+}$ by 2.1\% for early observations, and 4.2\% for very early observations. Similar results are obtained with LA-CODE, which even beats SSTH-RT$^{++}$ by a 2.6\% margin for low $\alpha$ and 4.1\% for very low $\alpha$. It surpasses SSTH-RT$^{+}$ by 4.1\% in the early $\alpha$ range and 6.1\% in the very early $\alpha$ range. Finally, it solidly outperforms SSTH-RT by 6.1\% and 9.4\% for early and very early observations, respectively.

We display mAP@K for LA-CODE for increasing observation levels, compared with SSTH-RT$^{++}$'s evaluated on \textit{complete} queries (SSTH-RT$^{++}_{\alpha=1}$) in Figure \ref{fig:mapk}. With 10\% of video observed at 192 bits, LA-RECO loses to SSTH-RT$^{++}_{\alpha=1}$ by 5\%, which is expected since it has access to far less information. However LA-RECO catches up with 30\% of video observed. At 80\% of video observed LA-RECO even surpasses SSTH-RT$^{++}_{\alpha=1}$ by a small margin. MAP@K plots for other bitcode sizes and methods are provided in the supplementary material.

\subsection{Qualitative results}
\label{qual}

We selected two FCVID queries from the query set on which LA-RECO returned successful results at the end of the video, i.e. the top 10 results are correct. We followed the evolution of the top 5 search results at different points in time. Both query videos depict the action \textit{"decorating a christmas tree"}. Retrieval might be subpar early on in the video, but the model should ideally adapt and return more relevant database entries as the video plays.

Figure \ref{fig:qual_1} shows an example of a successful case. A mistake is made early on at 20\%, but the top 5 search results improve to a perfect $5/5$ after 50\% of the video is seen. On the other hand, in Figure \ref{fig:qual_2} the model starts with $3/5$ correct results, then drops to $1/5$ in the middle. Retrieval eventually improves towards the end with $4/5$ correct results at 81\% progression. The video supplementary material (\url{https://youtu.be/Eq-lIUipd4E}) shows examples of \textbf{label-agnostic} similarity in the retrieved videos. For instance at 2:25 a video of a person operating a lawnmower is the query. \textbf{Certain results have mismatched labels such as cello or archery, while retaining the same fine-grained motion semantics, in these examples the arm movement.}

\section{Conclusion}

In this paper, we present an unsupervised video hashing approach designed for incremental retrieval performed mid-stream using incomplete video queries. To tackle this challenging new problem, we propose a predictive encoder that learns to generate enhanced bitcode representations of incomplete queries by acquiring knowledge distilled from another encoder trained on complete videos. Our approach yields a large performance improvement over the baseline. It also outperforms a naive approach that inflates the codebook by adding codes generated from truncated copies of the videos in the database, while being at the same time much faster and more scalable for live hashing and retrieval on video streams.
\paragraph{\textbf{Acknowledgements}} This work was supported by French state funds managed by the ANR under reference ANR-16-CE33-0009 (DeepSurg) and by BPI France (project CONDOR). The authors would also like to acknowledge the support of NVIDIA with the donation of a GPU used in this research.
\bibliographystyle{splncs}
\bibliography{egbib}
\end{document}